\documentclass[a4j]{article}

\usepackage{arxiv}

\usepackage[utf8]{inputenc} 
\usepackage[T1]{fontenc}    
\usepackage{hyperref}       
\usepackage{url}            
\usepackage{booktabs}       
\usepackage{amsfonts}       
\usepackage{nicefrac}       
\usepackage{microtype}      
\usepackage{lipsum}		

\usepackage{comment}
\usepackage[dvipdfmx]{color}

\usepackage[dvipdfmx]{graphicx}
\usepackage{cite}

\title{Ensemble learning in CNN augmented with fully connected subnetworks}

\author{
  Daiki Hirata\\
  Industrial Technology Center of Okayama Prefecture\\
  \texttt{daiki\_hirata@pref.okayama.lg.jp}
  \And
  Norikazu Takahashi\\
  Graduate School Natural Science and Technology, Okayama University\\
  \texttt{takahashi@cs.okayama-u.ac.jp}
  }

\begin{document}
\maketitle

\begin{abstract}
Convolutional Neural Networks (CNNs) have shown remarkable performance in general object recognition tasks. In this paper, we propose a new model called EnsNet which is composed of one base CNN and multiple Fully Connected SubNetworks (FCSNs). In this model, the set of feature-maps generated by the last convolutional layer in the base CNN is divided along channels into disjoint subsets, and these subsets are assigned to the FCSNs. Each of the FCSNs is trained independent of others so that it can predict the class label from the subset of the feature-maps assigned to it. The output of the overall model is determined by majority vote of the base CNN and the FCSNs. Experimental results using the MNIST, Fashion-MNIST and CIFAR-10 datasets show that the proposed approach further improves the performance of CNNs. In particular, an EnsNet achieves a state-of-the-art error rate of 0.16\% on MNIST. 
\end{abstract}


\section{Introduction}
Convolutional Neural Networks (CNNs) \cite{CNN} are attracting a great deal of attention because they show remarkable performance in general object recognition tasks. Various methods have been proposed so far for improving the performance of CNNs: pre-processing\cite{preprocess1, preprocess2, preprocess3}, dropout~\cite{dropout}, batch normalization~\cite{batchnorm}, ensemble learning \cite{Ensenble1, Ensemble2}, and so on. 

In this paper, we propose a new model based on CNNs to further improve the performance in image recognitioin tasks. Our model consists of one base CNN and multiple Fully Connected SubNetworks (FCSNs). The base CNN generates a set of multi-channel feature-maps after each convolutional layer. The set of feature-maps generated by the last convolutional layer is divided along channels into disjoint subsets, and each subset is assigned to one of the FCSNs, which is trained independent of others so that it can predict the class label from the subset of the feature-maps assigned to it. The output of the overall model is determined by majority vote of the base CNN and the FCSNs. Namely, ensemble learning is performed in the proposed method. We thus call this model {\em EnsNet} in this paper. It is known that, in order for ensemble learning to be effective, the base learners must represent certain degree of diversity. In the proposed model, it is expected that FCSNs have this property because different subnetworks are trained using different training data. 


In what follows, we first explain the architechture of the EnsNet and how to train it. We then provide results of some experiments using the MNIST\cite{MNIST}, Fashion-MNIST\cite{Fashion-MNIST}, and CIFAR-10\cite{Cifar-10} datasets, which show that the proposed approach certainly improves the performance of CNNs. In particular, it is shown that an EnsNet achieves a state-of-the-art error rate of 0.16\% on MNIST.

\begin{figure}[t]
\centering
\includegraphics[width=15cm,height=10cm]{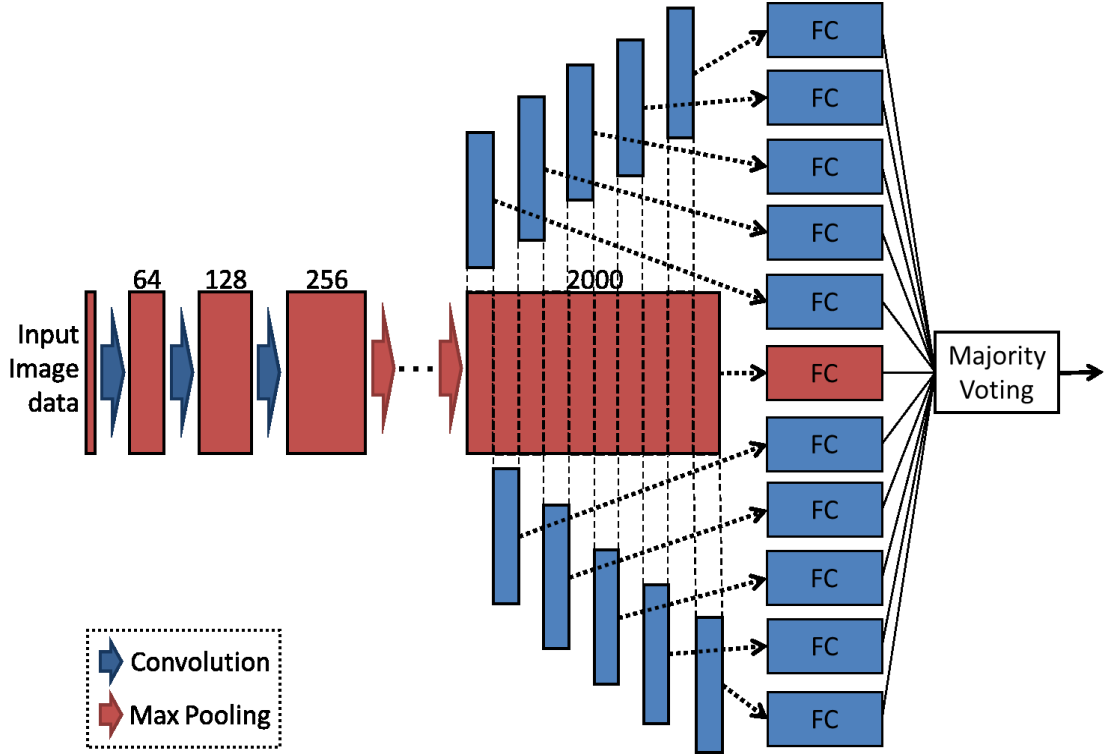}
\caption{An example of the architecture of the EnsNet. Red boxes and blue ones represent the base CNN and the subnetworks, respectively. The integer on the top of each red box is the number of channels of the feature-maps. 
}
\label{fig:proposed-model}
\end{figure}

\section{EnsNet}
\subsection{Architecture} \label{subsec:config}
The proposed model called EnsNet consists of one base CNN and multiple subnetworks as shown in Fig.~\ref{fig:proposed-model}. The structure of the base CNN varies depending on image recognition tasks. Table~\ref{tb:mainnet} shows two different structures used in the experiments shown in Section~\ref{sect:3}: one is for MNIST and Fashion-MNIST, and the other is for CIFAR-10. The ReLU activation function is used in the proposed model, though this is not explicitly shown in Table~\ref{tb:mainnet}. The set of feature-maps generated by the last convolutional layer of the base CNN is divided along channels into disjoint subsets, and each subset is fed into one of the subnetworks. Each subnetwork is a fully connected neural network consisting of multiple weight layers. Table~\ref{tb:subnet} shows the details of the structure of the subnetworks used in the experiments: one is for MNIST and Fashion-MNIST, and the other is for CIFAR-10. The output of the overall model is determined by majority vote of the base CNN and the subnetworks.


\begin{table}[t]
  \centering
    \caption{Structures of the base CNN used in the experiments. The left one is for the MNIST and Fashion-MNIST datasets, and the right one is for the CIFAR-10 dataset. Both CNNs have nine weight layers. The size of each convolutional layer is denoted as ``Conv<receptive field size>-<number of channels>'', and the size of each fully connected layer is denoted as ``FC-<number of nodes>''. The ReLU activation function is used in both models but is not shown in this table for simplicity. 
} 
    \label{tb:mainnet}
    \begin{tabular}[t]{|c|} \hline
      Input: $28\times28$ MNIST or Fashion-MNIST image\\ \hline
      Conv3-64 (zero padding)\\
      BatchNormalization\\
      Dropout(0.35)\\
      Conv3-128\\
      BatchNormalization\\
      Dropout(0.35)\\
      Conv3-256 (zero padding)\\
      BatchNormalization\\ \hline
      maxpool($2\times2$)\\ \hline
      Dropout(0.35)\\
      Conv3-512 (zero padding)\\
      BatchNormalization\\
      Dropout(0.35)\\
      Conv3-1024\\
      BatchNormalization\\
      Dropout(0.35)\\
      Conv3-2000 (zero padding)\\
      BatchNormalization\\ \hline
      maxpool($2\times2$)\\ \hline
      Dropout(0.35)\\ \hline\hline
      Dividing feature-maps (10 divition)\\ \hline\hline
      FC-512\\
      BatchNormalization\\
      Dropout(0.5)\\ \hline
      Dropconnect(0.5) \cite{Dropconnect, ChainerDropconnect} \\
      FC-512\\ \hline
      FC-10\\ \hline
      soft-max\\ \hline
    \end{tabular}
    \begin{tabular}[t]{|c|} \hline
      Input: $32\times32$ CIFAR-10 image\\ \hline
      Conv3-64 (zero padding)\\
      BatchNormalization\\
      Dropout(0.25)\\
      Conv3-128\\
      BatchNormalization\\
      Dropout(0.25)\\
      Conv3-256 (zero padding)\\
      BatchNormalization\\ \hline
      maxpool($2\times2$)\\ \hline
      Dropout(0.25)\\
      Conv3-512 (zero padding)\\
      BatchNormalization\\
      Dropout(0.25)\\
      Conv3-1024\\
      BatchNormalization\\
      Dropout(0.25)\\
      Conv3-2048 (zero padding)\\
      BatchNormalization\\ \hline
      maxpool($2\times2$)\\ \hline
      Dropout(0.25)\\
      Conv3-3000 (zero padding)\\
      BatchNormalization\\
      Dropout(0.25)\\
      Conv3-3500 (zero padding)\\
      BatchNormalization\\
      Dropout(0.25)\\
      Conv3-4000 (zero padding)\\
      BatchNormalization\\
      Dropout(0.25)\\ \hline\hline
      Dividing feature-maps (10 divition)\\ \hline\hline
      FC-512\\
      BatchNormalization\\
      Dropout(0.3)\\ \hline
      Dropconnect(0.3)\\
      FC-512\\ \hline
      FC-10\\ \hline
      soft-max\\ \hline
    \end{tabular}
\end{table}

\begin{table}[t]
  \centering
  \caption{The structures of the subnetworks used in the experiments. The left one is for MNIST and Fashion-MNIST datasets, and the right one is for CIFAR-10. All subnetworks have three weight layers.}
  \label{tb:subnet}
  \begin{tabular}[t]{|c|} \hline
      Input: $200\times6\times6$ feature-maps \\ (MNIST or Fashion-MNIST) \\ \hline
      FC-512\\
      BatchNormalization\\
      Dropout(0.5)\\ \hline
      Dropconnect(0.5)\\
      FC-512\\ \hline
      FC-10\\ \hline
      soft-max\\ \hline
  \end{tabular}
  \begin{tabular}[t]{|c|} \hline
      Input: $400\times7\times7$ feature-maps \\ (CIFAR-10)\\ \hline
      FC-512\\
      BatchNormalization\\
      Dropout(0.3)\\ \hline
      Dropconnect(0.3)\\
      FC-512\\ \hline
      FC-10\\ \hline
      soft-max\\ \hline
  \end{tabular}
\end{table}

\subsection{Training}
The EnsNet is trained by alternating two steps: one is the base CNN training step, and the other is the subnetworks training step. In the base CNN training step, the parameters of the convolutional layers and the fully connected layers of the base CNN are updated by some optimization algorithm, while the parameters of the subnetworks are fixed. In the experiments shown in Section~\ref{sect:3}, the Adam optimizer~\cite{Adam} is used. In the subnetworks training step, the parameters of the base CNN are fixed, and each subnetwork is trained independent of other subnetworks, using the corresponding subset of the feature-maps generated by the last convolutional layer of the base CNN and the target class labels as the training data. The parameters of the fully connected layers of each subnetwork are updated by the same optimization algorithm as the base CNN.

\begin{figure}[htp]
  \includegraphics[width=8cm]{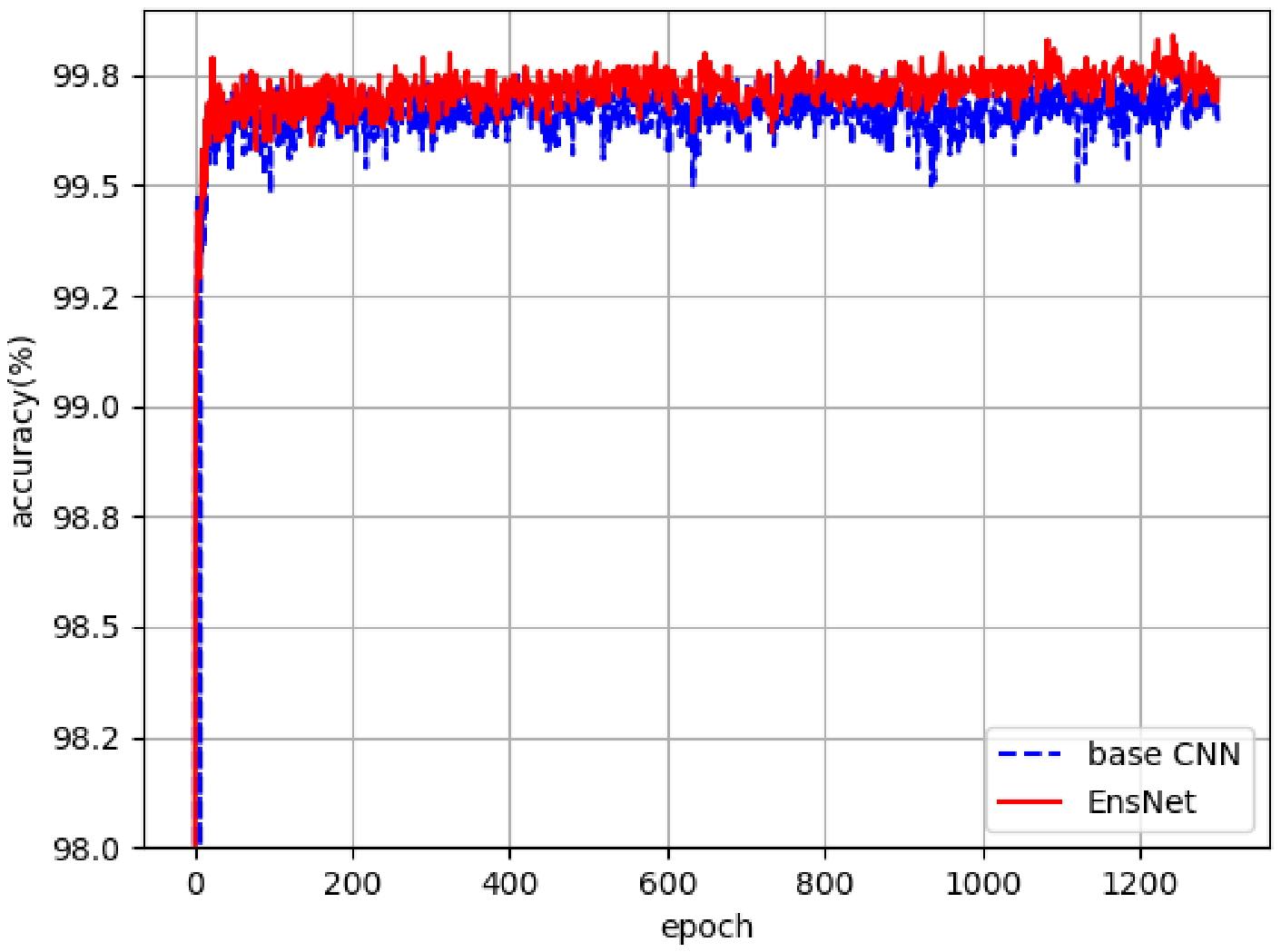}
  \includegraphics[width=8cm]{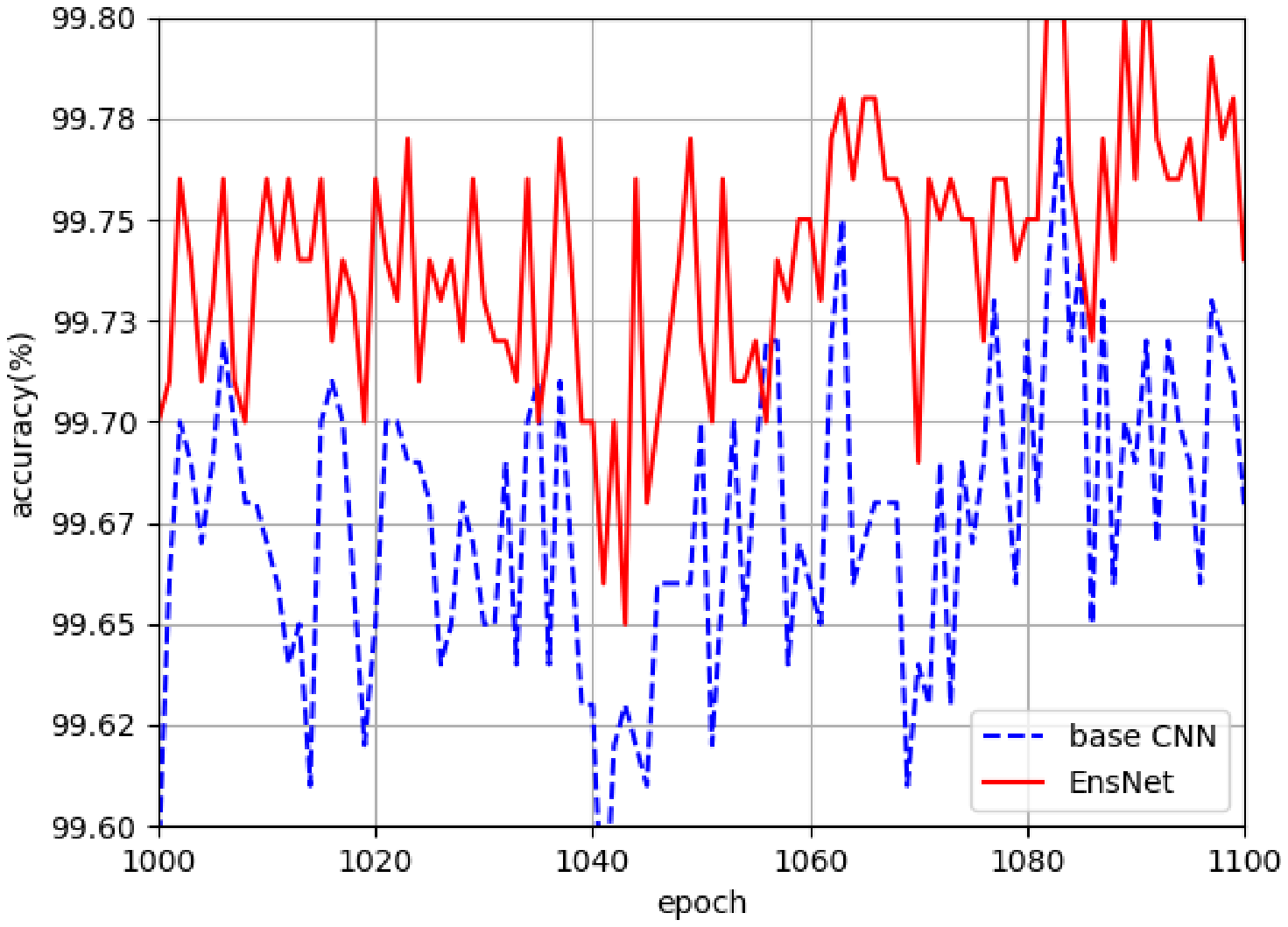}
  \caption{The test set accuracy of the EnsNet and the base CNN during training of the MNIST dataset (left) and a zoomed view of it (right). }
  \label{fig:improvement-by-DFM}
\end{figure}

\section{Classification Experiments}
\label{sect:3}
\subsection{Setup}
In order to evaluate the effectiveness of the EnsNet, we conducted classification experiments using the MNIST, Fashion-MNIST and CIFAR-10 datasets. The models used in the experiments were implemented in Chainer framework\cite{chainer}, and trained by the Adam optimizer~\cite{Adam}. The parameters of the Adam optimizer were set as follows: $\alpha = 0.001$, $\beta_1 = 0.9$, $\beta_2 = 0.999$, $\epsilon = 10^{-8}$, and a weight decay was set to $0$.

The MNIST dataset is a collection of $28 \times 28$ gray scale images of handwritten digits from $0$ to $9$. The training and test sets consist of 60,000 and 10,000 images, respectively. Before training, we augmented data by rotating images by various angles between $-10^\circ$ and $10^\circ$, scaling images by various factors between $0.8$ and $1.2$, shifting images to the width direction or the height direction by a fraction between $-0.08$ and $0.08$ of the total width or the total height, stretching images by the shear transformation with various angles between $-0.3^\circ$ and $0.3^\circ$. We also set the batch size and the number of epochs to 100 and 1,300, respectively. 

The Fashion-MNIST dataset is a collection of $28 \times 28$ gray scale images in 10 classes. 
The training and test sets consist of 60,000 and 10,000 images, respectively. Before training, we augmented data by rotating images by various angles between $-5^\circ$ and $5^\circ$. We also set the batch size and the number of epochs to 100 and 600, respectively. 

The CIFAR-10 dataset is a collection of $32 \times 32$ colored images in 10 classes. The training and test sets consist of 50,000 and 10,000 images, respectively. Before training, we augmented data by rotating images by various angles between $-10^\circ$ and $10^\circ$, scaling images by various factors between $0.8$ and $1.2$, shifting images to the width direction or the height direction by a fraction between $-0.08$ and $0.08$ of the total width or the total height, stretching images by the shear transformation with various angles between $-0.3^\circ$ and $0.3^\circ$. We also set the batch size and the number of epoch to 100 and 200. Furthermore, we decayed the learning rate $\alpha$ by $0.1$ every 100 epochs.

\subsection{Effect of Fully Connected Subnetworks}
In the first experiment, in order to evaluate the effectiveness of the fully connected subnetworks and the majority vote, we trained the EnsNet with the structure shown in the left columns of Tables~\ref{tb:mainnet} and \ref{tb:subnet}, and the base CNN in the EnsNet using the MNIST dataset, and measured how the test set accuracies of these two models change as the number of epochs increases. Fig.~\ref{fig:improvement-by-DFM} shows the results of this experiment. It is seen from Fig.~\ref{fig:improvement-by-DFM} that the test set accuracy of the EnsNet is higher than that of the base CNN. This means that the fully connected subnetworks and the majority vote can improve the performance of CNNs. 

\subsection{Comparison with Other Models}
In the second experiment, we trained the EnsNet and some conventional models including the base CNN using the MNIST, Fashion-MNIST and CIFAR-10 datasets, and measured the error rates for the test sets. Table~\ref{tb:prediction-error1} shows the error rates of six models: Random Multimodel Deep Learning for Classification (RMDL)~\cite{RMDL}, Dropconnect~\cite{Dropconnect}, Multi-Column Deep Neural Network (MCDNN)~\cite{MCDNN}, Augmented PAttern Classification (APAC)~\cite{APAC}, EnsNet, and the base CNN in the EnsNet for the MNIST dataset. Here, by EnsNet, we mean the model with the structure shown in the left columns of Tables~\ref{tb:mainnet} and \ref{tb:subnet}. The EnsNet acheived the error rate of 0.16\% which is lower than that of the RMDL, one of the state-of-the-art models for the MNIST dataset classification task. 


Table~\ref{tb:prediction-error2} shows the error rates of four models: Random Erasing~\cite{RandomErasing}, VGG8B(2x)+LocalLearning+CO~\cite{VGG8LCO}, EnsNet, and the base CNN in EnsNet for the Fashion-MNIST dataset. Here, by EnsNet, we mean the model with the structure shown in the left columns of Tables~\ref{tb:mainnet} and \ref{tb:subnet}. The EnsNet is not the best, but outperforms the base CNN. Table~\ref{tb:prediction-error3} shows the error rates of the EnsNet with the structure shown in the right columns of Tables~\ref{tb:mainnet} and \ref{tb:subnet}, and the base CNN in the EnsNet for the CIFAR-10 dataset. The EnsNet acheived a lower error rate than the base CNN. These results mean that the fully connected sunetworks and the majority vote certainly improved the performance of CNNs.

\begin{table}[t]
  \centering
  \caption{Test set error rates of six models for MNIST dataset.}
  \begin{tabular}[t]{|l|c|} \hline
    Model & Error rate\\
    \hline \hline
    RMDL~\cite{RMDL} & 0.18\%\\ \hline
    Dropconnect~\cite{Dropconnect} & 0.21\%\\ \hline
    MCDNN~\cite{MCDNN} & 0.23\%\\ \hline
    APAC~\cite{APAC} & 0.23\%\\ \hline
    EnsNet (Proposed) & \textbf{0.16}\% \\ \hline
    Base CNN in EnsNet & 0.21\% \\ \hline
  \end{tabular}
  \label{tb:prediction-error1}
\end{table}

\begin{table}[t]
  \centering
  \caption{Test set error rates of four models for Fashion-MNIST dataset.}
  \begin{tabular}[t]{|l|c|} \hline
    Model & Error rate \\
    \hline \hline
    Random Erasing~\cite{RandomErasing} & \textbf{3.65}\% \\ \hline
    VGG8B(2x)+LocalLearning+CO~\cite{VGG8LCO} & 4.14\% \\ \hline
    EnsNet (Proposed) & 4.70\% \\ \hline
    Base CNN in EnsNet & 5.00\%\\ \hline
  \end{tabular}
  \label{tb:prediction-error2}
\end{table}

\begin{table}[t]
  \centering
  \caption{Test set error rates of two models for CIFAR-10 dataset}
  \begin{tabular}{|l|c|} \hline
    Model & Error rate \\
    \hline \hline
    EnsNet (Proposed) & \textbf{23.75}\% \\ \hline
    Base CNN in EnsNet & 23.90\%\\ \hline
  \end{tabular}
  \label{tb:prediction-error3}
\end{table}

\section{Conclution}
We proposed a new CNN model called EnsNet, which is composed of one base CNN and multiple fully connected subnetworks. In this model, the set of feature-maps generated by the last convolutional layer of the base CNN is divided into disjoint subsets, and each subset is fed into one of the subnetworks as its input. The training of the EnsNet is done by updating the parameters of the base CNN and those of the subnetworks alternately, and the prediction is done by the majority vote of the base CNN and the subnetworks. Experimental results using the MNIST, Fashion-MNIST and CIFAR-10 datasets show that the EnsNet outperforms the base CNN. In particular, the EnsNet achieves the lowest error rate among some of the state-of-the-art models. A future work is to evaluate the effectiveness of our approach on other CNN models such as ResNet\cite{ResNet}.

\bibliographystyle{IEEEtran}
\bibliography{bibliography_revised_takahashi}

\end{document}